\documentclass[sigconf]{acmart}
\usepackage[cal=pxtx]{mathalfa}
\usepackage[ruled,vlined, linesnumbered]{algorithm2e}
\usepackage{pifont}

\newcommand{\xmark}{\text{\ding{55}}}
\usepackage{subfigure}
\usepackage{xcolor}
\usepackage{multirow}
\usepackage{float}
\definecolor{cadmiumgreen}{rgb}{0.0, 0.42, 0.24}
\AtBeginDocument{%
  \providecommand\BibTeX{{%
    \normalfont B\kern-0.5em{\scshape i\kern-0.25em b}\kern-0.8em\TeX}}}

\setcopyright{acmcopyright}
\copyrightyear{2021}
\acmYear{2021}

\acmConference[WWW'21]{Web Confernce 2021}{}

\begin{document}

\title{Stimuli-Sensitive Hawkes Processes for Personalized Student Procrastination Modeling
}
\author{Mengfan Yao}
\affiliation{
\institution{Department of Computer Science,  University at Albany - SUNY\\}}
\email{myao@albany.edu}

\author{Siqian Zhao}
\affiliation{
\institution{Department of Computer Science,  University at Albany - SUNY\\}}
\email{szhao2@albany.edu}
\author{Shaghayegh Sahebi}
\affiliation{
\institution{Department of Computer Science,  University at Albany - SUNY\\}}
\email{ ssahebi@albany.edu}
\author{Reza Feyzi Behnagh}
\affiliation{
\institution{Department of Educational Theory \& Practice,  University at Albany - SUNY\\}}
\email{rfeyzibehnagh@albany.edu}

\begin{abstract}

Student procrastination and cramming for deadlines are major challenges in online learning environments, with negative educational and well-being side effects. 
Modeling student activities in continuous time and predicting their next study time are important problems that can help in creating personalized timely interventions to mitigate these challenges. 
However, previous attempts on dynamic modeling of student procrastination suffer from major issues: they are unable to predict the next activity times, cannot deal with missing activity history, are not personalized, and disregard important course properties, such as assignment deadlines, that are essential in explaining the cramming behavior.
To resolve these problems, we introduce a new personalized stimuli-sensitive Hawkes process model (SSHP), by jointly modeling all student-assignment pairs and utilizing their similarities, to 
predict students' next activity times even when there are no historical observations. 
Unlike regular point processes that assume a constant external triggering effect from the environment, we model three dynamic types of external stimuli, according to assignment availabilities, assignment deadlines, and each student's time management habits.
Our experiments on two synthetic datasets and two real-world datasets show a superior performance of future activity prediction, comparing with state-of-the-art models. Moreover, we show that our model achieves a flexible and accurate parameterization of activity intensities in students.

\end{abstract}
\maketitle

\section{Introduction}
\label{sec:introduction}
Academic procrastination can be defined as postponing the planned studies, despite being aware of its negative consequences~\citep{Moon2005}.
This behavior is common in students, particularly in online education settings, in which students have to self-regulate their learning and studying ~\cite{Lee2011}.
Although there is no formal quantitative definition for procrastination, traces of this behavior can be observed by looking at students' study behavior, such as cramming the studies as deadlines approach ~\cite{Perrin2011}.
However, despite the negative side-effects of procrastination on students, such as on their academic performance and psychological well-being~\citep{Steel2007}, dynamic data-driven approaches that can model these indicator behaviors in students are scarce. 

Past research has mainly described student procrastination by summarizing student activities into static features~\cite{Perrin2011,Cerezo2017}, which cannot fully represent the dynamics of students' behavior through time.
More recently, sequential models of student behaviors have been used in the study of procrastination behaviors~\cite{Park2018,Yao2020}. 
However, these models fail to capture an important aspect of the cramming behavior: its relation to triggers such as course deadlines and availability of assignments, as in class schedule.
Additionally, these models are not personalized and do not model factors related to individual students, such as students' studying habits into account.
Finally, they cannot deal with missing activity data and fail to estimate students' next study times or predict their behavior in relation to various tasks and assignments.
An ideal student activity model should be able to capture students' response to the major events in the course, be personalized to learn student-specific behaviors, and be able to predict the students' future activity intensities as a way for early detection of procrastination, even if student sequence data is not completely observed.


Meanwhile, Hawkes processes~\cite{Hawkes1971}, as a family of Point processes, have shown great potential in dealing with complicated sequential data in most real-world applications, including in the education domain~\cite{Yao2020}. 
However, the state-of-the-art Hawkes process models used in the Education domain suffer from the above limitations, for two main reasons.
First, external stimuli and their triggering effects are conventionally parameterized as a constant, which results in ignoring factors such as class schedule and personalized student habits. 
For example, assignment deadline, as a format of external stimuli, may only start to show its triggering effect when it is approaching. 
Students' personal habits (e.g. log-in time and frequency) which is a reflection of their time management skills, also could evolve over time.
Secondly, the majority of the Hawkes processes model different sequences independently.
As a result, only future activities of the sequences with historical observations can be predicted, whereas the future of unobserved sequences can not be inferred. 

To address the above-mentioned limitations, we propose stimuli-sensitive Hawkes process (SSHP), that models the external course stimulus in addition to the internal activity stimulus, is personalized, and can predict the next activity times towards each assignment for students. 
In SSHP, we represent activities in each student-assignment pair as a Hawkes process.
To tackle the first aforementioned limitation, our model is designed to capture three types of external stimuli as parameterized functions of time: the effect of assignment availability, assignment deadline, and each student's personal study time and frequency habits. 
To deal with the second limitation, SSHP jointly models all student-assignment pairs, imposing a low-rank structure between student and assignment parameters in the model.
As a result, it can learn a personalized parameterization, even for unobserved sequences, based on the similarities shared between the students as well as the assignments.
Our extensive experiments on two synthetic datasets and two real-world datasets show a significant performance improvement in future activity predictions, compared with the state-of-the-art models, both when a sequence's data is partially or completely missing. 
We perform ablation studies on SSHP and show that all aspects of our model, including the external and internal parts, are important in contributing to its superior performance. 
And finally, we show the meaningful procrastination patterns that are captured by SSHP parameters, using clustering analysis and studying their associations with student performance in the course.

\section{Related Work}
\label{sec:related-work}
\noindent
\textbf{Procrastination Modeling in Education Domain.}
As there is no quantitative definition for procrastination behavior, in most of the recent educational data mining literature, procrastination-related behavior has been summarized by curating time-related features from student interactions in the course.
These studies aim to evaluate the relationships between these time-related features with student performance and do not model temporal aspects of procrastination~\cite{Baker2016,Cerezo2017,Kazerouni2017,agnihotriprocrastination}.
For example, Asarta et al. examined the students' log data from an online course use measures such as anti-cramming, pacing, completeness, etc.~\cite{asarta2013access}. However, such methods are static and can not describe students' varying behaviors over time.
For another example, Park et al. classify students into procrastinators and non-procrastinators by formulating a measure via a mixture model of per-day student activity counts during each week of the course~\cite{Park2018}. However, this is not able to model non-homogeneously spaced deadlines in a course.

As none of these models consider the timing of students' activities, they are not able to predict~\textit{when} the future activities will happen.
Sequential data modeling via point process could potentially deal with this limitation, however, it has not been applied to procrastination modeling until recently. 
To the best of our knowledge, the most related attempt that is comparable to ours has been made in~\cite{Yao2020}, where Yao et al. modeled each student's activities sequence as a Hawkes process and relates procrastination to the mutual excitation among activity types.
This work does not predict student's unseen activities, rather, a procrastination measure was proposed based on the learned parameters that have shown to be better correlated with students' grades than conventional delay measures.

\noindent
\textbf{Hawkes process and modeling scenarios.}
Hawkes processes, as a popular family of point processes, model two types of activities: activities that are triggered by external stimuli, and activities that are self-excited by the historical activities. 
The intensity of these two types of activities is usually parameterized by a base rate function and an excitation function, respectively.
To describe the complicated dynamics of real-world activity sequences, different state-of-the-art parameterizations have been proposed.
For example, Rizoiu et al. modeled the watching history of a Youtube video as a Hawkes process, and proposed to use the number of shares of a video on YouTube scaled by a constant to represent the base rate~\cite{rizoiu2017expecting}. 
In another example, Bao et al. proposed to use a sinusoidal function to capture the periodical rise-and-fall patterns of user activities on social media~\cite{bao2016modeling}. 
More recently, neural Hawkes models have been proposed to allow higher model capacity for learning arbitrary and nonlinear distributions of the history~\cite{du2016recurrent,mei2017neural,xiao2017modeling}. For example, Du et al. proposed to use RNN to model the arrival times of a given sequence and characterized the intensity as a function of the embedded hidden cell representations~\cite{du2016recurrent}.
Even though such neural-based Hawkes models allow for less bias and more flexibility than the traditional parametric models, they do not provide meaningful interpretations of the activity arrival patterns, which could be important to some scenarios such as procrastination analysis in educational settings.

In terms of the applications of sequential data modeling via Hawkes models, 
the majority of the state-of-the-art Hawkes models treat each individual sequence as an independent input, in other words, no relationship among the sequences is assumed. 
As a result, sequences without any observed history are usually excluded from the study.
To tackle this problem, a few state-of-the-art Hawkes process approaches model all sequences jointly by assuming underlying similarity among the sequences. 
For example, Du et al. modeled each user-product pair, i.e. the collection of interactions of a user to a product, as a Hawkes process. 
By assuming the similarity between users and products, their model learns the low rank representation of all Hawkes processes, including those that do not have historical purchasing history~\cite{Du2015}. 
Other similar approaches have also been proposed to measure sequences similarity by using auxiliary features~\cite{He2015,Li2018,Shang2018}. 
However, auxiliary information in education domains are usually excluded from the data due to privacy concerns.


\section{Stimuli-Sensitive Hawkes Process (SSHP)}
\label{sec:model}
\subsection{Problem Formulation}
\label{sec:problem}
Consider the case where there are $U$ students and $N$ assignments in a course. 
We assume that the time when student $u_i$ interacts with assignment $a_j$ depends on two things: (1) the effects of external stimuli (e.g., the deadline of $a_j$ is approaching, \textit{therefore} student $u_i$ starts to review the lectures and practices on the quizzes), and (2) the self-exciting nature of the events, in other words, past events can trigger the future ones (e.g., student $u_i$ decides to work on assignment $a_j$ \textit{because} they just watched the lecture video that is related to $a_j$).
To capture these triggering effects which can be important in explaining students behaviors in the course, we propose to model the collection of activity \textit{timestamps} of student $u_i$'s interactions with assignment $a_j$, or student-assignment pair $(u_i,a_j)$, as a point process (Sec.~\ref{sec:exp}.2), characterized by a function that captures the effects of both external stimuli and the effects of self-excitement (Sec.~\ref{sec:exp}.3). 
All the important notations used in the following section are summarized in Tbl.~\ref{tbl:notations}.
\begin{table}[]
\caption{A summary of important notations}\vspace{-10pt}
\resizebox{0.48\textwidth}{!}{%
\begin{tabular}{|l|l|l|l|}
\hline
\multicolumn{1}{|c|}{\multirow{7}{*}{Function}} &
  $\lambda$: intensity &
  \multirow{4}{*}{Vector} &
  $\mathbf{v} = (v_i)_N$ \\ \cline{2-2} \cline{4-4} 
\multicolumn{1}{|c|}{} & $f$: density                         &                      & $\mathbf{b} = (b_i)_N$               \\ \cline{2-2} \cline{4-4} 
\multicolumn{1}{|c|}{} & $\mu$: base rate                     &                      & $\mathbf{c} = (c_i)_N$               \\ \cline{2-2} \cline{4-4} 
\multicolumn{1}{|c|}{} & $s(t)$: self-excitement          &                      & $\mathbf{p} = (p_i)_N$               \\ \cline{2-4} 
\multicolumn{1}{|c|}{} &
  $\mathcal{L}$: loss &
  \multirow{3}{*}{Matrix} &
  $A = (\alpha_{ij})_{U\times N}$ \\ \cline{2-2} \cline{4-4} 
\multicolumn{1}{|c|}{} & $\mathcal{M}$: proximal operator &                      & $M = (m_{ij})_{U\times N}$           \\ \cline{2-2} \cline{4-4} 
\multicolumn{1}{|c|}{} & $P$: projection function         &                      & $\Gamma = (\gamma_{ij})_{U\times N}$ \\ \hline
\multirow{8}{*}{Scalar} &
  $\alpha$: self-exciting coef. &
  \multirow{4}{*}{\begin{tabular}[c]{@{}l@{}}Super-\\ script\end{tabular}} &
  $d$: deadline \\ \cline{2-2} \cline{4-4} 
                       & $\beta$: decay coef.              &                      & $h$: student habit                           \\ \cline{2-2} \cline{4-4} 
                       & $m$: deadline effects                   &                      & $o$: assignment opening                         \\ \cline{2-2} \cline{4-4} 
                       & $\gamma$: coef. in $\mu(t)$                 &                      & $S$: search point                    \\ \cline{2-4} 
                       & $v$: shape parameter             & \multirow{4}{*}{Set} & $X$: event sequence                        \\ \cline{2-2} \cline{4-4} 
                       & $p$: peak in $\mu^h$                  &                      & $\mathcal{O}$: observed sequences    \\ \cline{2-2} \cline{4-4} 
                       & $b$: base of $\mu^o$             &                      & $\Theta$: matrix parameter set       \\ \cline{2-2} \cline{4-4} 
                       & $c$: offset in $\mu^h$           &                      & $\phi$: vector parameter set         \\ \hline
\end{tabular}%
}
\label{tbl:notations}
\end{table}

\subsection{Modeling Student-Assignment Activity Timestamps}
Formally, given a student-assignment pair $(u_i,a_j)$, we describe it as the timestamps of all student $u_i$'s interactions with assignment $a_j$: $X{ij} = \{x_{ij}^{\tau}|\tau = 1,...,K_{ij}\}$
~\footnote{For simplicity, without causing any confusion, we omit the individual subscripts $i$ and $j$ in the rest of this section.}. 
Given time $t$, let $\mathcal{H}_t$ denote the historical observations in $X$ up to, but not including, time $t$, i.e. $\mathcal{H}_t = \{x^{\tau}|\tau = 1,...,n\}$, where $x^n$ is the time of the last event that took place before time $t$.
If the conditional p.d.f. (probability density function) of the next event's time is defined as $f^{*}:= f(t|\mathcal{H}_t)$,
the joint p.d.f. for a realization follows:
\begin{align}
\label{eq:pdf}
     f(x^1,...,x^{K})= \prod_{\tau=1}^Kf(x^{\tau}|\mathcal{H}_{\tau-1}) = 
    \prod_{\tau=1}^n f^{*}(x^{\tau}).
\end{align}
The above conditional p.d.f is one way to characterize a particular Hawkes process, however could be difficult for model design and interpretability~\cite{daley2007introduction}. 
Alternatively, in this work, we adopt a more commonly-used function for the characterization of Hawkes, i.e. the conditional intensity function, which can be shown to be a function of $f^{*}(t)$ and its corresponding cumulative distribution function $F^{*}(t)$:
\begin{align}
    \lambda(t) = \frac{f^*(t)}{1 -F^*(t)} 
    = \frac{f^*(t)}{1 - \int_{x^n}^t f^*(s|\mathcal{H}_{x^n})ds}
\end{align}
\subsection{Parameterization of External Stimuli and Self-excitement}
As mentioned above, we assume that there are two types of activities in the sequence of a given student-assignment pair, i.e., activities that are excited by the external stimuli, and those are self-excited by the previous activities. The intensities of both types of activities are respectively parameterized by a base rate function and a excitation function, defined as follows:

\noindent
\textbf{Modeling external stimuli.}
We parameterize the following $3$ types of external stimuli that can trigger student's interactions with the assignment.
Firstly, the effect of student habit: 
we assume that each student interacts with the course based on their own periodical studying schedule. 
For example, some students habitually log in the course at noon every day, but some prefer to study after midnight.
Secondly, the decaying effect of the assignment availability (opening): 
we assume that students activities can be triggered once the assignment is posted.
However, this effect decays over time.
For example, once an assignment is posted, students may log in and check the assignment requirements or deadlines, or revisit it later for the detailed descriptions. However, over time, this effect will die out and will be dominated by other stimuli.
Finally, the deadline of an assignment:  
we assume that student activities can be triggered by the deadline, and this effect gets stronger by approaching the deadline and wears off eventually. 

Formally, we define the base rate intensity for students at each time $t$ as a combination of each of the above stimulus as in Equation~\ref{eq:mu}.
\begin{align}
        \label{eq:mu}
    \mu(t) &= \gamma^d\mu^d(t) + \gamma^o\mu^o(t) + \gamma^h\mu^h(t),\\
    \label{eq:mut}
    \mu^h(t) & = \sin(\frac{2\pi}{s} (t + p) ) + c,\\
    \label{eq:muo}
    \mu^o(t) &= b^{t/s},\\
    \label{eq:mud}
    \mu^d(t)  &= 
    \begin{cases} 
   \frac{1}{\sqrt{2\pi v}(d - m-t/s)}e^{-\frac{(\ln{(d - m -t/s))^2}}{v}} & \text{if $d - m \leq t/s$,} \\
0    & \text{if $d - m > t/s$}.
\end{cases}
\end{align}


Specifically,
Eq.~\ref{eq:mut} models the activity intensity triggered by students \textit{\underline{h}abit} as a sinusoidal function. In other word, $\mu^h(t)$ captures periodicity of length $s$, that peaks at $p$. $c$
can be interpreted as the minimum number of the activities triggered by the student habits, which works as a base of $\mu^h(t)$.
Eq.~\ref{eq:muo} models the \textit{\underline{o}pening} effects of the assignment as an exponential function parameterized by $b$, with a decay speed of $1/b$ over time scaled by $s$.
This formulation will result in exponentially less number of activities, as a result of assignment posting, as time passes.
Eq.~\ref{eq:mud} models the effect of \textit{\underline{d}eadline} via a reversed log-normal function. $d$ here is the known time of the assignment deadline,
$d - m$ represents the time when the deadline's triggering effect on student activities is over.
As a result, $m$ represents the difference between the end of the deadline's effect and the deadline. 
If the effect of deadline is over after the actual time of deadline (e.g. late submission), $m$ would be negative. Otherwise, $m\geq 0$.
Non-negative $v$ controls how intense the activities are closing to the deadline and how fast this effect decays after the peak. 
This formulation represents that student activity intensities will peak around their last assignment-related activity, which is close to the deadline, either before or after it.
$\gamma^h$, $\gamma^o$ and $\gamma^d$ respectively are the weight coefficients that describe the importance of $\mu^h(t)$, $\mu^o(t)$ and $\mu^d(t)$.

\noindent
\textbf{Modeling internal stimuli.}
To model the effect of past activities, we adopt the following conventional self-excitation function used in point processes:
\begin{align}
    s(t) = \sum_{x^{\tau} <t }\alpha\beta e^{-\beta(t - x^{\tau})},
\end{align}    

The above excitation function characterizes the effect of each historical event $x^\tau$ to current time $t$, as a decaying function of the time difference between $t$ and $x^\tau$, with the decaying speed of $1/\beta$. Therefore, the more recent a historical event is, the more effect it has in terms of self-excitation.
$\alpha$ can be shown to be the branching ratio under this definition, i.e. the expected number of activities that are triggered by a given activity. Thus it is called self-exciting coefficient.

\noindent
\textbf{Intensity function.}
Finally, our intensity function for one student-assignment pair can be defined as follows :
\begin{align}
\label{eq:intensity}
    \lambda(t) &= \mu(t) + s(t)\\\nonumber
    &= \gamma^h(\sin(\frac{2\pi}{s} (t + p))+c) + \gamma^ob^{t/s} \\\nonumber
    & + \gamma^d (\frac{1}{\sqrt{2\pi v}(d -m-t/s)}e^{-\frac{(\ln{(d - m -t/s))^2}}{v}}) + \sum_{x^{\tau} <t }\alpha\beta e^{-\beta(t - x^{\tau})}.
\end{align}
As we can see, the intensity is the combination of base rate function $\mu(t)$ that models external stimuli, and the excitation function $s(t)$ that models the self-excitement.
The proposed intensity function falls in the category of a popular family of point process, i.e. Hawkes processes, which conventionally model the effect of all external stimuli as a constant.
As our proposed model parameterizes the effects of different external stimuli in educational setting as functions of time, we call our model Stimuli-Sensitive Hawkes process model (SSHP).

\noindent
\textbf{Matrix representation for all student-assignment pairs.}
Equation~\ref{eq:intensity} above represents the intensity function for activities of one student on one assignment.
To model all student activities on all assignments, one can model them as separate sequences and learn the parameters for each sequence independently. 
However, this kind of model will result in two limitations.
Firstly, no parameters can be learned for student-assignment sequences that are completely unobserved, and thus, student activities in such sequences cannot be predicted. 
For example, consider a student, who has not started working on a future assignment by the end of the observation window, or a student, who skips an assignment for now and plans to come back to it later. 
Excluding these sequences from the study largely limits the capacity of the model in our application. 
Secondly, the parameters of the model that are not assignment-related, such as student habit parameters, are going to be learned independently for each sequence. As a result, they will lose meaning.
A common approach to deal with these limitations is to extend the data collection window, which could be costly and inefficient. 
Another solution could be using the learned parameters from the observed sequences and applying them to the sequences that do not have observations. 
However, such an approach cannot provide personalized inferences, thus is not ideal.

To deal with these problems, while learning personalized parameters for students, we assume similarity between the learned parameters for all student-assignment pairs.  
Particularly, we represent the relationship between students and assignments as a student-assignment matrix, where a row is a student and each column represents an assignment from the course.
We represent the student-assignment related parameters of the model in such a matrix format, model the student-related parameters of the model in a vector format (so that they are shared between all assignments for a student), and share some generic parameters of the model between all students. 
As a result, for example, the intensity function of student-assignment pair $(u_i,a_j)$ can be defined as the parameters correspond to the $j$-th cell in row $i$ from the parameter matrices. 
More specifically, the parameters are set to follow the following three structures:

(1) \textit{ scalars:} following the convention of Hawkes processes, we set global decay coefficient $\beta$ to be shared among all sequences. We also set $s$ to be a global scalar, so that time $t$ is scaled to the same unit across all student-assignment pairs. 

(2) \textit{vector sets $\phi$:} We let $\mathbf{c}$ = $(c_1,...,c_U)$, $\mathbf{p} = (p_1,...,p_U)$, $\mathbf{b} = (b_1,...,b_U)$ and $\mathbf{v} = (v_1,...,v_U)$ to be vectors, assuming a student's habit is unchanged across the assignments (i.e. $c$ and $p$). Similarly, their sensitivity to the effect of assignment openings (i.e. $b$). 
Furthermore, how fast their activities becoming intense once the deadline started affecting them (i.e. $v$) is also set to be shared among assignments. 

(3) \textit{low-rank matrices $\Theta$:} For each of the rest of the parameters, we consider a matrix format and assume similarity among student-assignment pairs, i.e. a low rank structure on the matrix format. 


\subsection{Objective Function}
\textbf{Maximum likelihood estimation on one sequence.}
Given a student assignment pair $(u_i,a_j)$'s historical activities $X_{ij} = \{x_{ij}^{\tau}|\tau = 1,...,K_{ij}\}$ over the time period $[0,T]$, and a parameter set $\theta = (\alpha,\beta,s, p, c, b, v, m, \gamma^h, \gamma^o, \gamma^d)$, the likelihood $L$ is the joint probability of observing all historical events till time $T$, which has the following form~\cite{daley2007introduction}:
\begin{align}
\label{eq:gen_likelihood}
    L(X;\theta) = \prod_{\tau =1}^{K} f^*(x^\tau)= \prod_{\tau =1}^{K}\lambda(x^{\tau}) \cdot e^{(-\int_0^T\lambda(u)d(u))},
\end{align}
where $f^*(t)$ and $\lambda(t)$ are respectively the p.d.f defined in Eq.~\ref{eq:pdf} and the intensity function in Eq.~\ref{eq:intensity}.
Directly taking the log of the above equation to obtain the log-likelihood entails $\mathcal{O}(K^2)$ complexity due to the double summations - 
i.e. the summation in Eq.~\ref{eq:intensity} combined with the summation term introduced by the log of the product from Eq.~\ref{eq:gen_likelihood}. To achieve a more feasible complexity of $\mathcal{O}(K)$, we use the recursive function $R(\cdot)$ defined as follows:
\begin{align}
    R(\tau)  = 
\begin{cases} 
\big(1+R(\tau -1)\big)e^{-\beta\big(x^{\tau}-x^{\tau-1})}\big) & \text{if $\tau>1$,} \\
0 & \text{if $\tau =1$}.
\end{cases}
\end{align}
As a result, the final explicit form of log-likelihood $l(\theta)$ can be shown as below:
\begin{align}
\label{eq:likelihood}
    l(X;\theta) &= \log L(\theta)  = \sum_{\tau=1}^K\log (\lambda(x^{\tau}) - \int_0^{x^K}\lambda(u)du\\\nonumber
    &=\sum_{\tau} \log\big(\gamma^d\mu^d(x^{\tau})+\gamma^o\mu^o(x^{\tau})+\gamma^h\mu^h(x^{\tau})\\\nonumber
    &+\alpha\beta R(\tau)\big)- \gamma^dU^{d*}(x^{K}) - \gamma^o U^{o*}(x^{K}) - \gamma^h U^{h*}(x^{k}) \\\nonumber
    &+ \alpha\sum_{\tau=1}^K(e^{-\beta(x^{K}-x^{\tau})} -1 ).
\end{align}
$U^{d*}(\cdot)$, $U^{o*}(\cdot)$, $U^{h*}(\cdot)$ is respectively the cumulative intensity of $\mu^d$, $\mu^o$ and $\mu^h$ introduced due to the integral in Eq.~\ref{eq:gen_likelihood}, which can be obtained as below:
\begin{align}
    U^{h*}(x^{\tau}) & = \int_0^{x^{\tau}}\mu^h(u) du \\\nonumber
                &=\frac{1}{\pi}\big(-24s\cos(\frac{\pi x^{\tau} + \pi ps}{24s}) + 24s\cos\frac{\pi p}{24} + \pi c x^{\tau}\big),\\
        U^{o*}(x^{\tau}) & = \int_0^{x^{\tau}}\mu^o(u) du
                = s(\frac{b^{x^{\tau}/s }-1}{\ln{b}}),\\
    U^{d*}(x^{\tau}) & = \int_0^{x^{\tau}}\mu^d(u) du
                = -\dfrac{s\left(\operatorname{erf}\left(\frac{\ln\left(-\frac{x^{\tau}-(d-m)s}{s}\right)}{\sqrt{v}}\right)-\operatorname{erf}\left(\frac{\ln\left(d-m\right)}{\sqrt{v}}\right)\right)}{2^\frac{3}{2}}.
\end{align}
where $\text{erf}(x) = \frac{2}{\sqrt{\pi}}\int_0^x e^{-t^2} dt$ is the Gauss error function. 

\noindent
\textbf{Matrix representation of all sequences.}
Thus far, one could model a single student-assignment pair via SSHP based on its historical observations by maximizing the log-likelihood function defined in Eq.~\ref{eq:likelihood}.
However, as mentioned in the previous section, we represent some of the parameters ($\Theta$) in a matrix format for all student-assignment pairs and assume similarity among them, i.e. a low rank structure on the matrix. 
Specifically, we denote the set of the vector parameters as $\phi = \{\mathbf{c},\mathbf{p},\mathbf{b},\mathbf{v}\}$, and the set of the matrix parameters as $\Theta = (A,M,\Gamma^h,\Gamma^o,\Gamma^d)$ and impose a low-rank structure on all $\Theta$ in our objective function.
By using trace norm as a surrogate for low-rank structure, we constraint the trace-norm of $A = (\alpha_{ij})_{U\times N}$, $M = (m_{ij})_{U\times N}$, $\Gamma^h = (\gamma^h_{ij})_{U\times N}$, $\Gamma^o = (\gamma^o_{ij})_{U\times N}$, and $\Gamma^d = (\gamma^d_{ij})_{U\times N}$ to be small. 
\noindent
\textbf{Loss for all sequences.}
Finally, we can formulate the objective function as follows, based on the collection of observed sequences $\mathcal{O}= \{X_{ij}, ~\text{s.t.}~|X_{ij}|>0\}$:
\begin{align}
    \min_{\Theta,\phi}\mathcal{L} = & -\frac{1}{|\mathcal{O}|}\sum_{X_{ij}\in\mathcal{O}} l(\mathcal{X}_{ij}; \Theta_{ij},\phi_{i}) \\\nonumber
    \text{s.t. }& \mathbf{A} \geq 0, \mathbf{\Gamma}_d \geq 0, \mathbf{\Gamma}_o \geq 0, \mathbf{\Gamma}_h\geq0,  \mathbf{c}\geq 1, \mathbf{v}>0, 1>\mathbf{b}>0\\\nonumber
    & tr(\theta_u)\leq k_u,~\text{for}~\theta_u\in\Theta_{ij}.
\end{align}
The main objective is the negative log-likelihood of observing all sequences with events, while the non-negative constraint on $A$ is introduced to fit the definition of Hawkes that the sequences are self-exciting. 
All coefficients of the 3 types of external stimuli are also set to be non-negative.
$\mathbf{c}$ is constrained to be greater than or equal to $1$ to make sure the non-negative effect of student habit with the use of sinusoidal function, and each $\mathbf{v}$ element is the shape parameter in the reversed log-normal function thus needs to be positive.
Each cell of $\mathbf{b}$ is set to be constrained between $0$ and $1$ to meet the assumption that the effect of assignment opening is decaying but not increasing or unchanged. 
We also constrain each parameter $\theta_u$'s trace norm in the matrix format to be small, which is equivalent to constraining the rank of $\theta_u$ to be less than or equal to $k_u$.

\subsection{Parameter Inference}
We adopt Accelerated Gradient Method (AGM)~\cite{nesterov2013gradient} framework for the inference of parameters.
Our choice is for having a faster convergence rate, especially when we have both non-smooth trace norm and non-negativity in the constraints. 
The key subroutines of AGM in our model can be summarized as follows. 
For a matrix format parameter $\theta_u\in\Theta$, the objective is to compute the proximal operator:
\begin{align}
    \theta_u^* &= \text{argmin}_{\theta_u}\mathcal{M}_{\gamma,\theta_u^S}(\theta_u) \\\nonumber
    &= \text{argmin}_{\theta_u}\frac{\gamma}{2}\|\theta_u - P_{\theta_u}(\theta^S_u - \frac{1}{\gamma}\nabla_{\theta_u}\mathcal{L})\|_F^2.
\end{align}
$\gamma$ is the step size, $\theta^S_u$ is used to denote the current search point  of $\theta_u$, and $\nabla_{\theta_u}\mathcal{L}$ is the gradient of loss $\mathcal{L}$ w.r.t $\theta_u$. 
$P_{\theta_u}(\cdot)$ is a projection function to make sure the parameter value at each step is properly constrained. 
More specifically, $P_{\theta_u}(\cdot)$ for all $\theta_u\in\Theta$ is set to be $\big(TraceProj(\cdot)\big)_{+}$ where the inner $TraceProj(\cdot)$ is a trace projection~\cite{cai2010singular} and the outer $(\cdot)_+$ projects negative values to $0$.
Similarly, the key subroutine for the inference of $\phi_u\in\phi$ is shown as follows:
\begin{align}
    \phi_u^* &= \text{argmin}_{\phi_u}\mathcal{M}_{\phi_u^S,\gamma}(\phi_u) \\\nonumber
    &= \text{argmin}_{\phi_u}\frac{\gamma}{2}\|\phi_u - P_{\phi_u}(\phi_u^S - \frac{1}{\gamma}\nabla_{\phi_u}\mathcal{L})\|_F^2.
\end{align}
$P_{\phi_u}(\cdot)$ is also a projection function that makes sure the constraint of $\phi_u$ is met. When a value falls out of the constrained interval, it is projected to the closet value within the interval.

We also present Algorithm~\ref{algo} to effectively solve the objective according to the subroutines mentioned above.

\begin{algorithm}[h]
\small
\SetAlgoLined
\SetKwInput{KwOutput}{Output}  
\KwIn{$\eta > 1$, step size $\gamma_0$, MaxIter}
initialization: $\theta_{u,1} = \theta_{u,0}~\text{for}~\theta_{u}\in\Theta$; $\phi_{u,1} = \phi_{u,0}$ for $\phi_{u}\in\phi$; $\alpha_{0} = 0; \alpha_1 = 1$\;
\For{$i = 1$ to MaxIter}{
  $a_i = \frac{\alpha_{i - 1} -1}{\alpha_{i}}$\;
  update $\theta_{u,i}^S = \theta_{u,i} + a_i(\theta_{u,i} - \theta_{u,i-1})$ for all $\theta_u\in\Theta$\;
  update $\phi_{u,i}^S = \phi_{u,i} + a_i(\phi_{u,i} - \phi_{u,i-1})$ for all $\phi_u\in\phi$\;

     \While{Ture}{
    compute $\theta_u^* = P_{\theta_u} (\theta_{u,i}^S-\nabla \mathcal{L}/\gamma_i )$~\text{for all}~$\theta_u\in\Theta$\;
    compute $\phi_u^* =  P_{\phi_u}(\phi_{u,i}^S-\nabla \mathcal{L}/\gamma_i)$~\text{for all}~$\phi_u\in\phi$\;
      \eIf{$\mathcal{L}(X;\Theta^*,\theta^*_i) \leq \mathcal{L}(X;\Theta^S,\phi^S) + \sum_{\theta\in\{\Theta,\phi\}}\langle \theta^S_i,\nabla \mathcal{L}(\theta^S_i)\rangle +  \alpha_k/2\|\theta^S_i - \theta^*_i\|_F^2$}{
       break\;
       }{
       $\gamma_i = \gamma_{i-1} \times \eta$\;
      }
     
  $\theta_{u,i+1} = \theta^{*}$ for all $\theta_u\in\Theta$\;$\phi_{u,i+1} = \phi^{*}$ for all $\phi_u\in\phi$ \;
  \eIf{\text{stopping criterion satisfied}}
  {break\;}
  {$\alpha_i = \frac{1 + \sqrt{1 + 4 \alpha_{i-1}^2}}{2}$}
     }
     }
\KwOutput{$\theta_u = \theta_{u,i+1} ~\text{for all}~\theta_u\in\Theta,  \phi_u = \phi_{u,i+1}~\text{for all} ~\phi_u\in\phi$}
 \caption{Algorithm for parameter inference of SSHP}
 \label{algo}
\end{algorithm}

\section{Experiment Setup and Baselines}
\label{sec:expset}
In this section, we first introduce the state-of-the-art approaches that we used as baselines in Sec.~\ref{sec:expset}.1. An introduction of both synthetic and real-world datasets is given in Sec.~\ref{sec:expset}.2. Finally, the experiment setup including train-test splitting and hyperparameters tuning is presented in Sec.~\ref{sec:expset}.3.
\subsection{Baseline Approaches}
In this work, we compare the proposed SSHP to the following $7$ baselines considering different aspects, i.e., model parameterization, modeling strategy (if can generate personalized predictions for unobserved sequences), and application scenarios.

\noindent\textbf{Poisson~\cite{kingman2005p}}: We use the Poisson process model as the simplest baseline, where the intensity function is characterized by the event arrival rate. 

\noindent\textbf{HRPF~\cite{hosseini2018recurrent}}: The state-of-the-art Poisson factorization model proposed by Hosseini et al. (among the proposed models in the paper, this is the version that does not require user-network as auxiliary features). 
All sequences are modeled jointly, therefore, unobserved sequences can be predicted as well.

\noindent\textbf{RMTPP~\cite{du2016recurrent}}: A state-of-the-art Neural Hawkes model that uses RNN to model the dependencies between past and future events in a sequence. The intensity function of this Hawkes model is defined based on the hidden states. All sequences are assumed to be independent.

\noindent\textbf{ERPP~\cite{xiao2017modeling}}: Another state-of-the-art Neural Hawkes model, which models auxiliary features as time series. These time series and the event sequences are modeled by two separate LSTMs. Similar to RMTPP, all sequences are modeled independently.

\noindent\textbf{DHPR~\cite{hosseini2018recurrent}}: A variation of HRPF, where an excitation parameter is used to capture the self-excitement in the sequences. However, the excitation is represented as a hyperparameter that is shared among all sequences.

\noindent\textbf{HPLR~\cite{Du2015}}: The state-of-the-art user-item recommendation model using Hawkes processes. This model can be seen as an improvement of DHPR, in which the excitation parameter can be learned for all sequences. 

\noindent\textbf{EdMPH~\cite{Yao2020}}: The most recent approach that studies student procrastination using Hawkes processes. All activities of a student during the course are modeled in a sequence, independent from other sequences. 

A summary of the baselines is presented in Table~\ref{tbl:baselines}.

\begin{table}[]
\caption{A summary of baseline approach and the proposed SSHP}\vspace{-15pt}
\resizebox{\linewidth}{!}{%
\begin{tabular}{|c|c|c|c|c|}
\hline
Model & Self-exciting & \begin{tabular}[c]{@{}c@{}}Non-constant\\ base of time\end{tabular} & \begin{tabular}[c]{@{}c@{}}Infer completely \\ missing seq.\end{tabular} & \begin{tabular}[c]{@{}c@{}}Application\\ in Education\end{tabular} \\ \hline
Poisson & $\color{red}{\xmark}$       & $\color{red}{\xmark}$       & $\color{red}{\xmark}$       & $\color{red}{\xmark}$       \\ \hline
HRPF    & $\color{red}{\xmark}$       & $\color{red}{\xmark}$       & $\color{cadmiumgreen}{\checkmark}$ & $\color{red}{\xmark}$       \\ \hline
RMTPP   & $\color{cadmiumgreen}{\checkmark}$ & $\color{cadmiumgreen}{\checkmark}$ & $\color{red}{\xmark}$       & $\color{red}{\xmark}$       \\ \hline
ERPP    & $\color{cadmiumgreen}{\checkmark}$ & $\color{cadmiumgreen}{\checkmark}$ & $\color{red}{\xmark}$       & $\color{red}{\xmark}$       \\ \hline
DHPR    & $\color{cadmiumgreen}{\checkmark}$ & $\color{red}{\xmark}$       & $\color{cadmiumgreen}{\checkmark}$ & $\color{red}{\xmark}$       \\ \hline
HPLR    & $\color{cadmiumgreen}{\checkmark}$ & $\color{red}{\xmark}$       & $\color{cadmiumgreen}{\checkmark}$ & $\color{red}{\xmark}$       \\ \hline
EdMPH   & $\color{cadmiumgreen}{\checkmark}$ & $\color{red}{\xmark}$       & $\color{red}{\xmark}$ & $\color{cadmiumgreen}{\checkmark}$ \\ \hline\hline
\textbf{SSHP}    & $\color{cadmiumgreen}{\checkmark}$ & $\color{cadmiumgreen}{\checkmark}$ & $\color{cadmiumgreen}{\checkmark}$ & $\color{cadmiumgreen}{\checkmark}$ \\ \hline
\end{tabular}%
}
\label{tbl:baselines}
\vspace{-10pt}
\end{table}

\subsection{Datasets}
\begin{figure}[!ht]
    \centering
        \caption{Activity intensity of a sampled student-assignment pair.}
    \includegraphics[width = \linewidth]{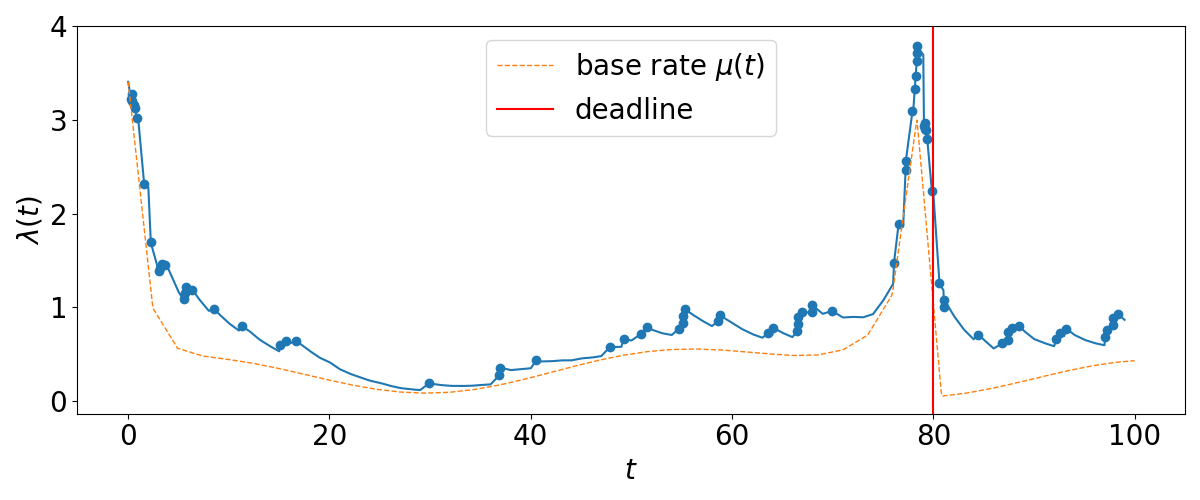}
    \label{fig:sampled}
    \vspace{-15pt}
\end{figure}

\textbf{Synthetic Data. }
Presuming $500$ students and $20$ assignments, we created $10k$ $(500\times20)$ simulated student-assignment pairs, and sampled $\sim 100$ events for each pair using the Ogata thinning algorithm~\cite{ogata1988statistical}, which is the most commonly used sampling method in the related literature. 
Specifically, we used the intensity function defined in Eq.~\ref{eq:intensity} and sampled each of its parameters from normal distributions, where $A\sim\mathcal{N}(0.4,0.1)$,  $M\sim\mathcal{N}(0,5)$, $\Gamma^d\sim\mathcal{N}(15,3)$, $\Gamma^o\sim\mathcal{N}(5,3)$,  $\Gamma^h\sim\mathcal{N}(0.5,0.1)$, $\mathbf{v}\sim\mathcal{N}(20,10)$,  $\mathbf{b}\sim\mathcal{N}(0.5,0.3)$, $\mathbf{p}\sim\mathcal{N}(6,4)$, and $\mathbf{c}\sim\mathcal{N}(1.2,0.1)$.
We empirically set these distributions to approximate the intensity patterns observed in real data.
For visualization, Fig.~\ref{fig:sampled} shows a sequence generated by open library tick~\cite{2017arXiv170703003B}, in which all the parameters are set to be the means. 
The solid blue line shows the sequence intensity, where each blue dot represents a sampled activity, the dashed orange line is the base rate, and the synthetic deadline is $80$, shown as the vertical red line.
To simulate the real-world scenarios, in which only some of the data sequences can be observed, we created two datasets, randomly masking $10\%$ (named as Syn-$10$ data) and $90\%$ (named as Syn-$90$ data) of the sequences to be unobserved.
In other words, $10\%$ of the sequences from Syn-$10$ dataset, and $90\%$ of the sequences from Syn-$90$ dataset are unobserved.

\noindent
\textbf{Computer Science Course on Canvas Network (CANVAS). } 
This real-world dataset is from the Canvas Network online platform~\cite{Canvas-Network2016}  
that hosts various open courses in different academic disciplines. 
The computer science course we use happens during $\sim6$ weeks. 
In each week, a graded assignment-style quiz is published in the course resulting in $6$ graded course assignments. 
From this dataset, we obtain $\sim729$K timestamps of $384$ student-assignment pairs.
Activities include submission activities, module learning (reading, watching videos, etc.) activities, and discussions.

\noindent
\textbf{Big Data in Education on Coursera (MORF)} Our second real-world dataset 
is collected from an 8-week ``Big Data in Education'' course on the Coursera platform. 
The dataset is available through the MOOC Replication Framework (MORF)~\cite{andres2016replicating}. 
In total, we extract $\sim52$K activities from $246$ students-assignment pairs, that contain quiz and assignment activities, watching lecture videos, and discussion-related activities.

\subsection{Experiment Setup}
We test our method in two scenarios according to our application: 
1) when the historical observations are available, we want to predict what will happen in the future based on the history,
and 2) when the whole sequence of activities for a student-assignment pair is completely missing, we want to infer its future without observing its history.
To test the model's performance in predicting the future in these two scenarios, we split our data into the following $3$ sets: \textbf{training set} that contains the initial historical observations, which is used to train the model for parameter inference; 
\textbf{partially missing test set} that contains the rest of the historical observations, that is used for testing the first scenario.
Finally, the \textbf{completely missing test set} contains the entire observations of the sequences, and it is used to examine models' ability in generating personalized and accurate predictions for unobserved sequences, i.e. scenario 2.  

For Syn-$10$, we naturally set the $10\%$ masked sequences to be the completely missing test set.
In the remaining 90\% unmasked sequences, we use the first $70\%$ of the activities (i.e. synthetic past observations) to be training and the later $30\%$ (i.e. synthetic future activities to be predicted) to be partial missing testing. 
We perform a similar procedure on Syn-$90$, with 90\% masked sequences to be completely missing test set, and a $70\%-30\%$ split in the remaining sequences for training and partially missing testing respectively.
For both real-world datasets, we randomly holdout $20\%$ of the sequences to be completely missing, and for the rest of the $80\%$ sequences, we also use the same $70\%-30\%$ split to generate training and partially missing testing.

For the baseline models that are not able to generate personalized predictions of future times without historical observations (i.e. Poisson, RMTPP, ERPP, and EdMPH), we report the root mean squared error (RMSE) of the time prediction on the partially missing test set only, and for the other models, we report the RMSE on both partially and completely missing test sets. 

The hyperparameters of proposed SSHP across all datasets are tuned via grid search on the following values:
global decay $\in \{1,6,12\}$; initial step size $\gamma_0$ $\in \{1,10,100,200\}$; update speed $\eta \in \{2,5,10,20\}$; and trace norm penalty in trace norm projection $rho \in \{0.01,0.1,1\}$. 
For the synthetic datasets, the best hyperparemters are set as follows: we have decay $\beta = 1$, the step size $\gamma_0 = 100$, update speed is $\eta = 2$, trace norm penalty is $1$. In CANVAS, decay $\beta = 6$, the step size $\gamma_0 = 200$, $\eta = 5$, trace norm penalty is $1$. In MORF, we have decay $\beta = 1$, $\gamma_0 = 200$, $\eta = 2$, and $\rho = 0.1$.
Similarly, the hyperparameters for baseline approaches are tuned via grid search according to the ranges provided in the original papers.

\section{Fit and Arrival Time Prediction}
\label{sec:exp}

In the following set of experiments, we study SSHP's ability to recover the correct parameters for the underlying processes, investigate its performance in predicting the next activity time compared to the state-of-the-art baselines, and analyze the contribution of different parts of the model in its performance.


\subsection{Model Fit on Synthetic Data}
As a way to evaluate SSHP's performance in capturing the sequence dynamics, we investigate its ability to find the true parameters of the underlying processes.
Since these parameters are available from the synthetic datasets, we calculate the root mean squared error (RMSE) between the estimated parameter values by SSHP and the actual parameter values that have been used to generate the synthetic datasets.
The results are shown in Tbl.~\ref{tbl:sim-pars}.
Generally, SSHP performs better in the partially missing test set than in the compleletly missing test.
That is because the task of learning completely unobserved sequences without histories is more challenging than learning sequences with partially observed histories.
Additionally, the results show that the RMSEs in Syn-$90$ dataset are only marginally higher than in Syn-$10$ in both partially and completely missing test sets.
This suggests the model's robustness and its potential to recover the parameters even when the ratio of unobserved sequences is high in the dataset.

\begin{table}[!ht]
\resizebox{0.49\textwidth}{!}{%
\begin{tabular}{|l|l|c|c|c|c|c|c|c|c|c|}
\hline
\multicolumn{2}{|l|}{Datasets}                          & $\mathbf{v}$ & $\mathbf{b}$ & $\mathbf{p}$ & $\mathbf{c}$ & $A$  & $M$  & $\Gamma^d$ & $\Gamma^o$ & $\Gamma^h$ \\ \hline
\multirow{2}{*}{Syn-10} & part. miss. & 1.33 & 0.1  & 1.33 & 0.09 & 0.05 & 2.64 & 1.65 & 1.08 & 0.16 \\ \cline{2-11} 
                        & compl. miss.  & 1.23 & 0.12 & 1.39 & 0.16 & 0.13 & 2.60 & 2    & 1.54 & 0.13 \\ \hline
\multicolumn{1}{|c|}{\multirow{2}{*}{Syn-90}} & part. miss. & 1.34         & 0.10         & 1.33         & 0.09         & 0.06 & 2.39 & 1.80       & 1.14       & 0.18       \\ \cline{2-11} 
\multicolumn{1}{|c|}{}  & compl. miss. & 1.31 & 0.12 & 1.38 & 0.16 & 0.12 & 2.61 & 1.97 & 1.51 & 0.17 \\ \hline
\end{tabular}%
}
\caption{RMSE of parameters learned by proposed model SSHP in synthetic datasets.}
\label{tbl:sim-pars}
\end{table}

\begin{figure}[!ht]
    \centering
    \includegraphics[width =0.48\textwidth]{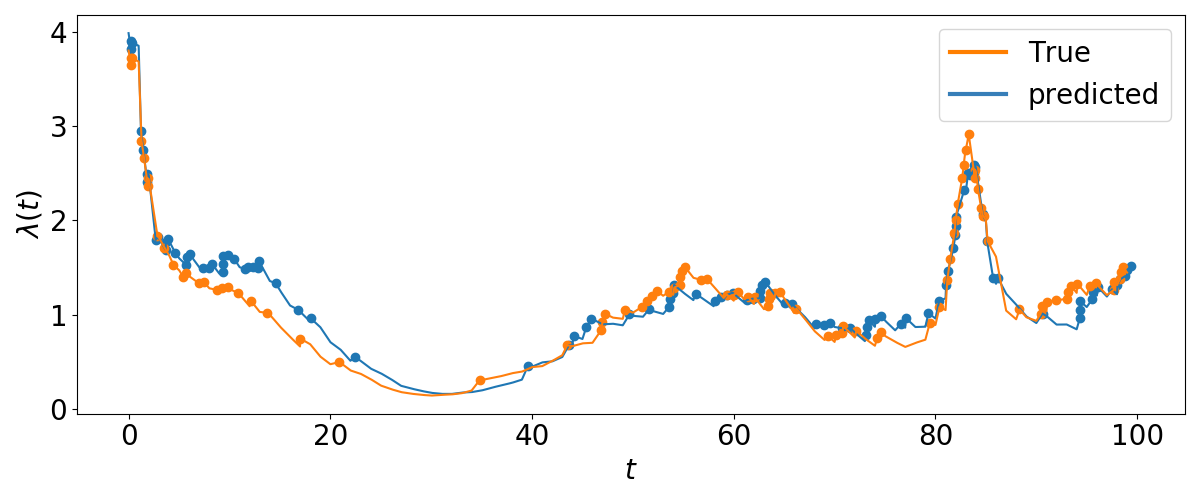}
    \caption{Predicted Intensity of a synthetic sequence.}
    \label{fig:sim-intensity}
    \vspace{-15pt}
\end{figure}

To provide a visual representation of these results, Fig.~\ref{fig:sim-intensity} shows the sampled intensity of a real sequence in Syn-$90$ dataset and the predicted intensity that is sampled based on the predicted parameters.
This figure demonstrates the model's ability in accurately capturing the dynamics of the sequence.

\subsection{Predicting Future Event Arrival Times}
Predicting the arrival times of future events for a given sequence, is the most commonly used evaluation method in the related literature. 
More formally, for a student-assignment pair, the arrival time of future $z$-th event after observation window, denoted as $x_{z}$, can be computed as the expectation of the sequence intensity w.r.t to time $t$. 
However, since time is continuous and the intensity functions of Hawkes processes are usually complicated, the analytic form of this expectation is hard to obtain.

Alternatively, in this work, we adopt another popular approach to predict future event arrival times.
We first use Ogatha thinning algorithm to sample inter-arrival times $\Delta t_z$, which is the time difference between $z$-th and $(z-1)$-th events.
Then, we compute the predicted time of $z$-th event $\hat{x}_{z}$ 
as $\hat{x}_{z} = \hat{x}_{z-1} + \frac{1}{N_t}\sum_{i=1}^{N_t} \Delta t^i_z$, 
where $N_t$ is the trail number for the sampling and $\Delta t^i_z$ is the sampled inter-arrival time at the $i$-th trail. 
The intuition is that inter-arrival times are sampled $N_t$ times, then the sample mean of all $N_t$ trails is used as the approximation of the actual inter-arrival time. 
In this way, we can recursively sample the arrival times for future events from the last historical observation and the learned intensity function.

In this work, we evaluate the model performances in predicting the next $10$ future activities after the observation window is ended, using RMSE between the actual and predicted times as our measure.
As the number of future activities grows, the task of predicting their arrival times becomes more challenging.

\begin{figure}[!ht]
\centering
\subfigure{
\includegraphics[width=0.48\textwidth]{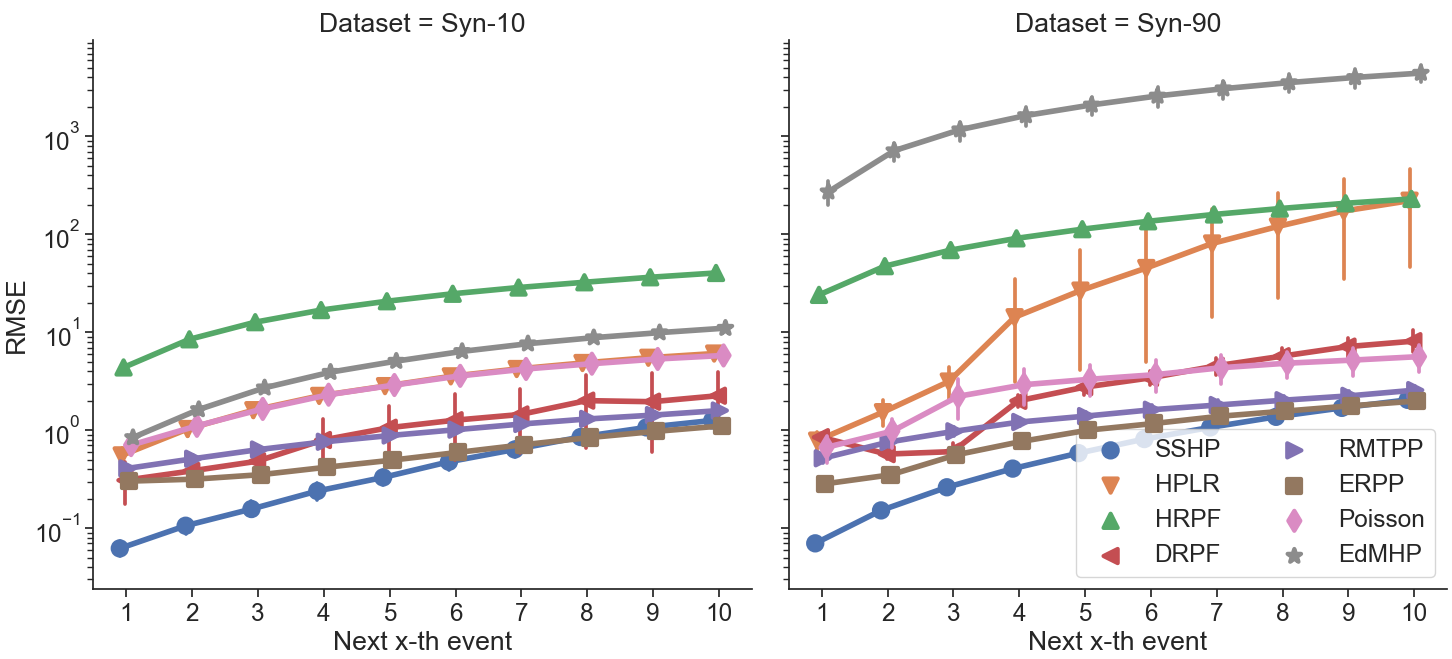}
}
\caption{Time prediction RMSE on \textit{partially missing test set} set with 95\% confidence interval on Syn-$10$ and Syn-$90$. }
\label{fig:simu-predict-test}

\subfigure{
\includegraphics[width=0.48\textwidth]{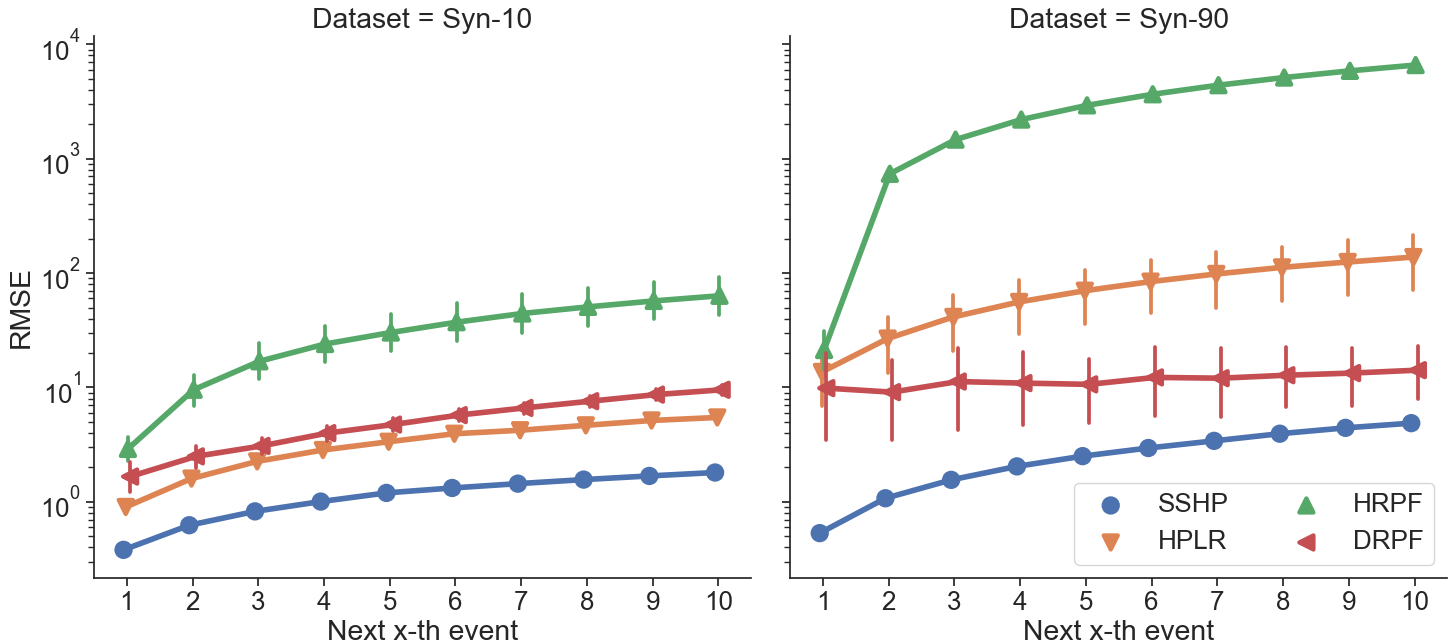}
}
\caption{Time prediction RMSE on \textit{completely missing test set} with 95\% confidence interval on Syn-$10$ and Syn-$90$. }
\label{fig:simu-predict-holdout}
\end{figure}
\noindent\textbf{Performance on synthetic datasets.}
In this section, we present the experiment results for SSHP and baseline approaches on synthetic datasets.
Fig.~\ref{fig:simu-predict-test} shows the model performances in partially missing test set in both Syn-$10$ and Syn-$90$, while Fig.~\ref{fig:simu-predict-holdout} shows the performances in the completely missing test set. 
The x-axis represents the future events' indices. 
For example, $x = 2$ represents the second event in the future after the end of the observation period $T$.
The y-axis is RMSE of time predictions in the log-scale, for a clearer separation between the models in the figures.
Some baselines, such as ERPP and RMTPP, are missing from the lower plots since they cannot predict unobserved sequences  (student-assignment sequences in completely missing test set).
We can see that SSHP clearly achieves the smallest RMSE of time predictions comparing to the baseline approaches in all settings.
Even though neural models ERPP and RMTPP start to show better performances in later event predictions, they are not able to predict unobserved sequences (i.e. completely missing test set).
As expected, since recovering completely unobserved sequences is more challenging, SSHP's performance on the partially missing test set is better than its performance in the completely missing test set set.

\begin{figure}[!t]
\centering
\subfigure{
\includegraphics[width=0.48\textwidth]{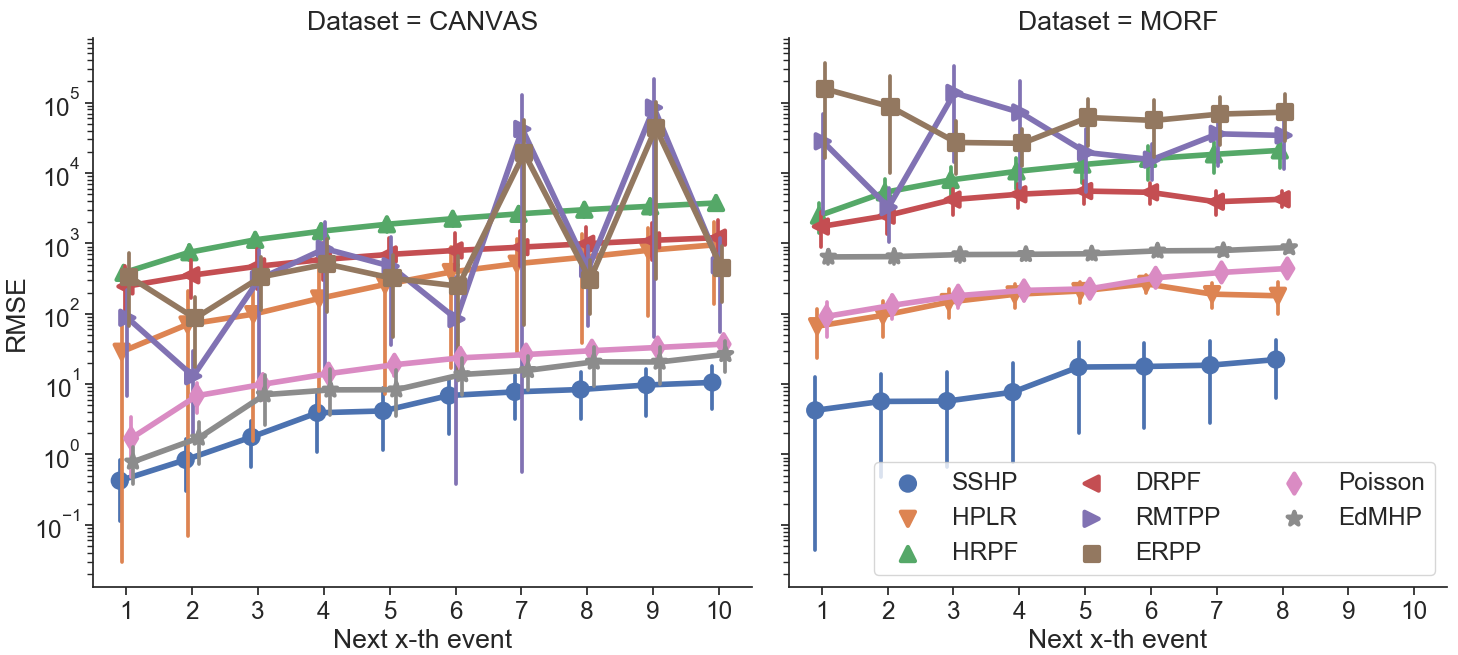}
}
\caption{Time prediction RMSE on \textit{partially missing test set} with 95\% confidence interval on two real-world datasets. }
\label{fig:real-predict-test}
\subfigure{
\includegraphics[width=0.48\textwidth]{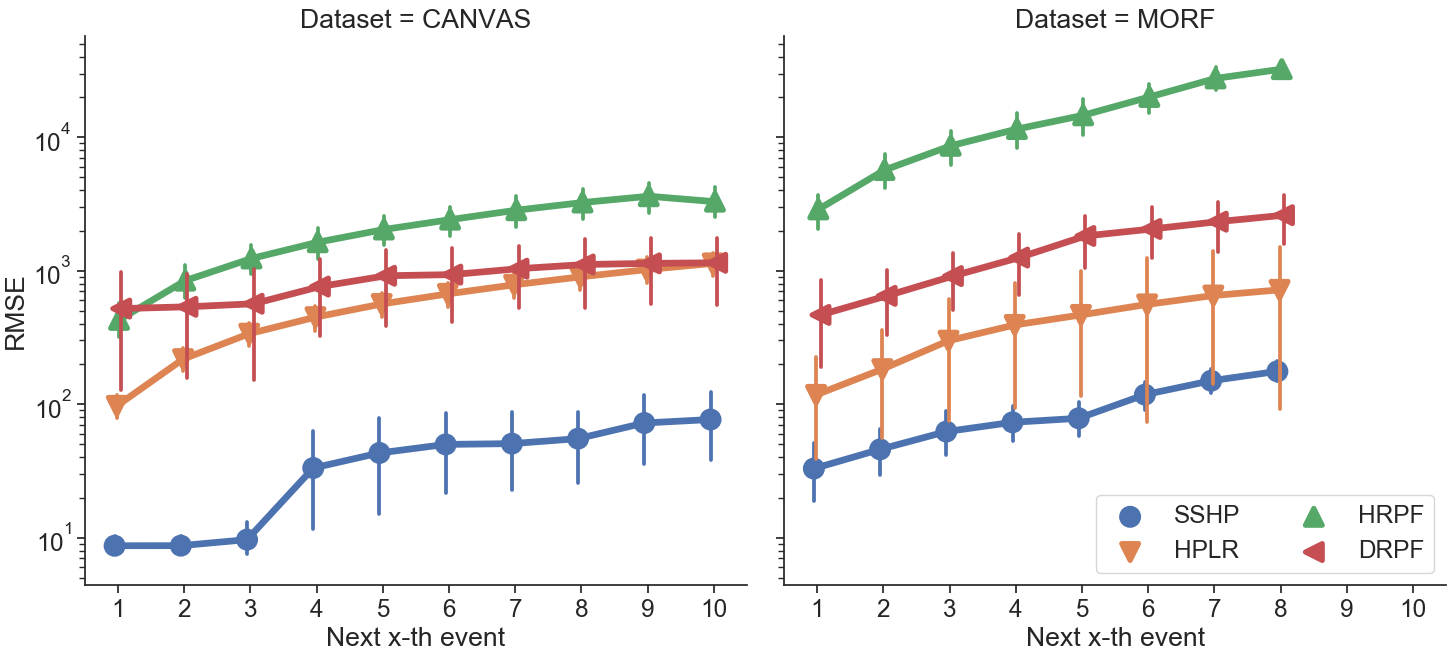}\hspace{-5pt}
}
\caption{Time prediction RMSE on \textit{completely missing test set} with 95\% confidence interval on two real-world datasets. }
\label{fig:real-predict-holdout}
\end{figure}

\noindent\textbf{Performance on real-world datasets.}
Next, we evaluate each model's performance using the two real-world datasets.
It is worth mentioning that the observed history in MORF is the shortest among all datasets, having an average of less than $\sim 26$ observations per sequence, and $\sim 18$ observations for training. For this reason, the prediction window is set to be $8$ in MORF instead of $10$ to achieve meaningful evaluation.
The evaluation in partially missing test set and completely missing test set is respectively presented in Fig.~\ref{fig:real-predict-test} and Fig.~\ref{fig:real-predict-holdout}.
As it is shown in the figures 
the proposed SSHP model outperforms the baseline approaches, especially by a big margin in Canvas's completely missing test set.
This is consistent with the synthetic dataset results.
In contrast, the performances of neural models ERPP and RMTPP are not as promising as they are in the synthetic datasets, especially in MORF.
One possible explanation is that the short training sequences in MORF restrict the ability of neural based models.

Another observation is that for higher indexed events in MORF's completely missing set and for lower indexed events in Canvas's partially missing test set, we observe overlapped confidence intervals between HPLR and SSHP, suggesting a less siginificant difference between two models' performances.
However, as the large confidence interval in HPLR suggests, its results are not robust and vary too much in the experiments. 
A potential explanation for the good predictions in HPLR is that for some student-assignment pairs the activity dynamics are rather invariant and a constant base rate, as in HPLR, is sufficient to capture them.

In conclusion, SSHP has shown to have superior time prediction performance in both synthetic and real-world datasets comparing with baseline approaches, especially on the challenging task of predicting the future for the completely missing test set. 

\subsection{Ablation Study}
To verify each component's importance in the intensity function, we compare SSHP to its variations SSHP-$s$, SSHP-$o$, SSHP-$h$ and SSHP-$d$, which respectively represents the model achieved by taking out the following components: 
self-excitement $s(t)$, effect of assignment opening $\mu^o(t)$, effect of student habit $\mu^h(t)$ and effect of deadline $\mu^d(t)$. 

\begin{figure}[!h]
\centering

\subfigure{
\includegraphics[width=0.48\textwidth]{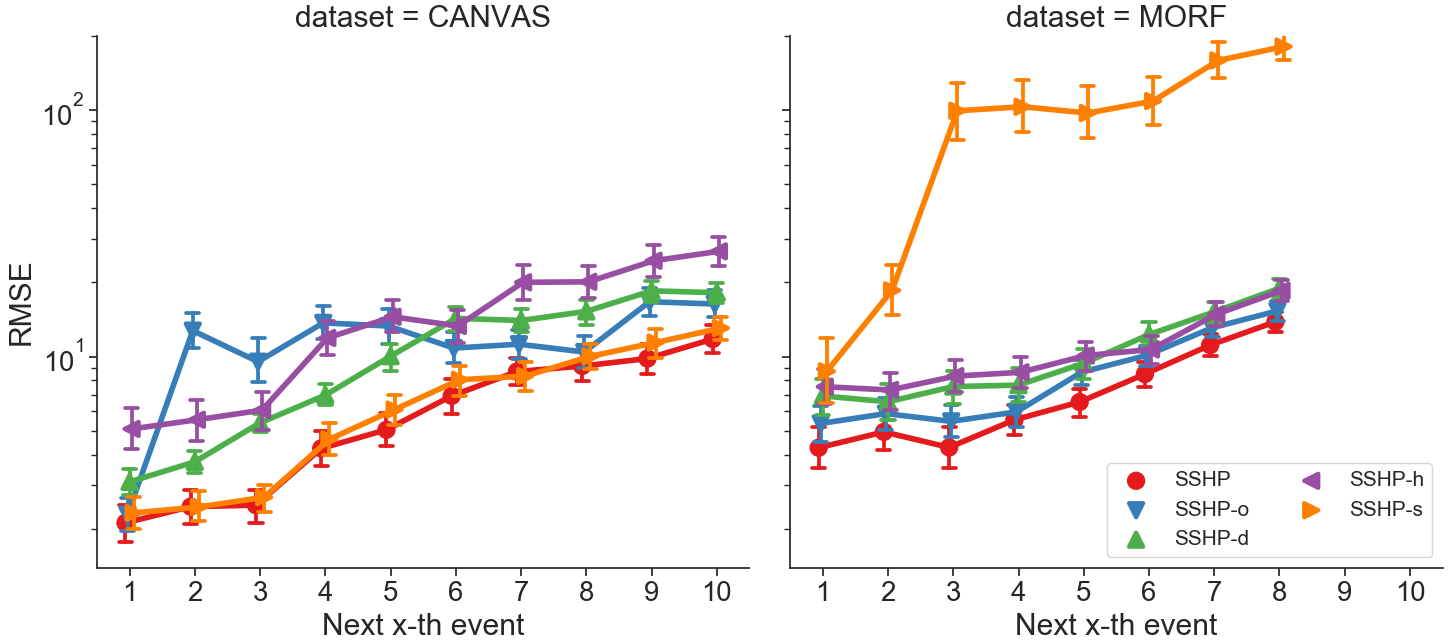}
}
\caption{Time prediction RMSE with 95\% confidence interval of SSHP and variations on \textit{partially missing test set}.}\label{fig:ablation-test}
 \hspace{-10pt}
\subfigure{
\includegraphics[width=0.48\textwidth]{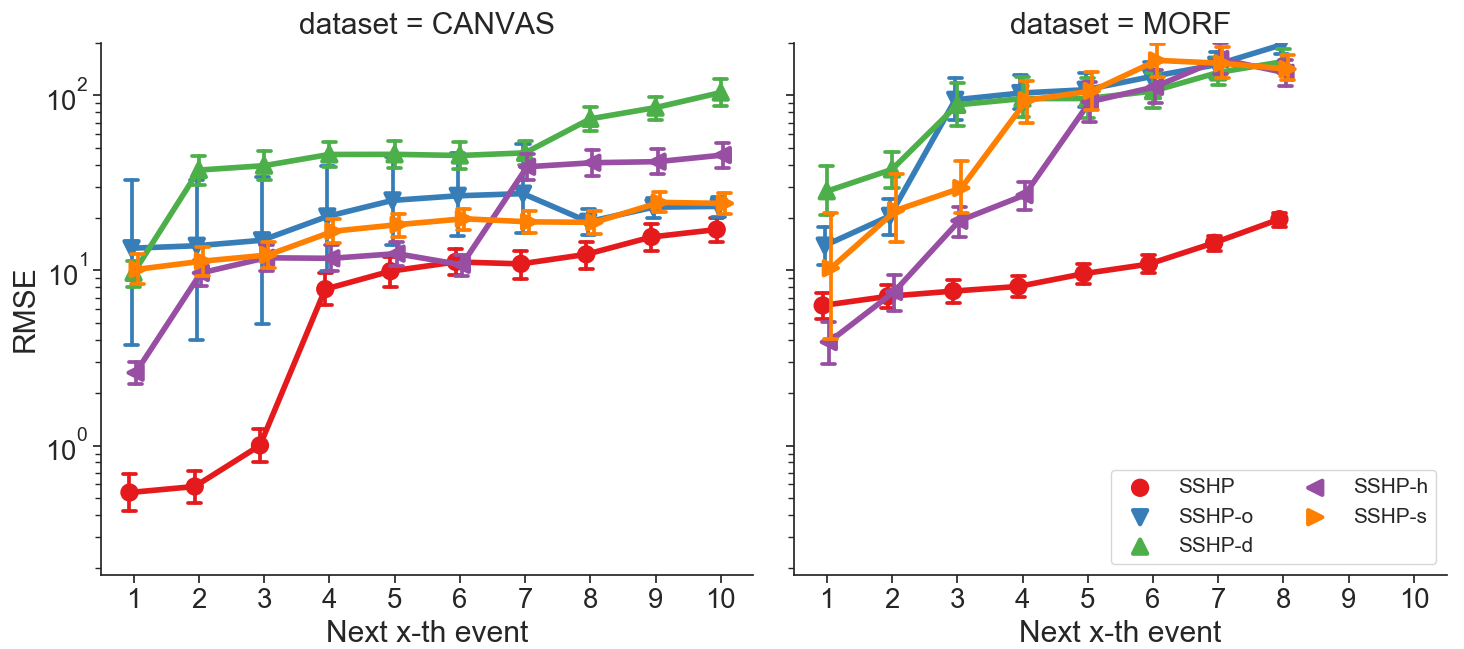}
}
\caption{Time prediction RMSE with 95\% confidence interval of SSHP and variations on \textit{completely missing test set}.}\label{fig:ablation-holdout}

\vspace{-10pt}
\end{figure}

Fig.~\ref{fig:ablation-test} and Fig.~\ref{fig:ablation-holdout} show the performance of these models in comparison with each other and with SSHP in respectively partially and completely missing test set.

In general, SSHP achieves lower time prediction errors in both real-world datasets, indicating the importance of each individual component. 
Furthermore, in the partially missing test set as shown in Fig.~\ref{fig:ablation-test}, while the improvement of modeling self-excitement is only marginal comparing with SSHP in CANVAS (left figure), self-excitement is shown to be a major factor in MORF (right figure), as SSHP-$s$ has higher prediction errors in MORF than other variations.

Additionally, as shown in Fig.~\ref{fig:ablation-holdout}, we can see the differences between SSHP and its variations are much more distinct in the completely missing test set (i.e. when the history is unobserved).
More specifically, in CANVAS dataset (left figure), we see that SHPP-$d$'s error is the highest among all models. 
This shows strong evidence of the deadlines' effect on student activities in CANVAS, which also suggests the importance of modeling $\mu^d(t)$. 
On the other hand in MORF, when comparing SSHP and SSHP-$h$ (right figure), we can see that the effect of student habit is not presented at the beginning of the sequence, as the error is lower when this stimulus is not included.
However, the importance of including student habits in the model is significant after the second event.
Another interesting observation is the higher confidence interval presented in SSHP-$o$.
One explanation is that some students are more sensitive to assignment opening compared to the others, therefore excluding $\mu^o(t)$ from the equation can cause a higher error to some sequences but not the others. 
The difference that is observed in the components' importance in the two datasets can come from the different nature of the two educational systems and the presented courses.
For example, one expects the effect of the deadline to be more prevalent in courses with a high late-submission penalty, compared to the ones with a more flexible scheme.

To conclude, despite the different characteristics that have been unveiled in the two datasets, we can see that all three external stimuli and the self-excitement components are important in modeling student activities. 

\section{Procrastination Pattern Discovery}
\label{sec:analysis}


In Section.~\ref{sec:model}, we have described the intuition behind SSHP's parameterization.
In this section, we analyze these parameters to demonstrate their interpretation and their association with student performance patterns.

\subsection{Cluster Analysis}
First, we investigate if the learned parameters can describe students' behaviors in assignments in a meaningful way that shows their cramming and procrastination behaviors.
To do so, we cluster
all student-assignment pairs via K-Means clustering algorithm, representing each of student-assignment item as its learned parameters: $(u_i,a_j) = (\alpha_{ij}, m_{ij},\gamma^d_{ij},\gamma^o_{ij},\gamma^h_{ij},v_i, b_i,p_i,c_i)$.

\begin{figure}[!ht]
    \centering
    \includegraphics[width=\linewidth]{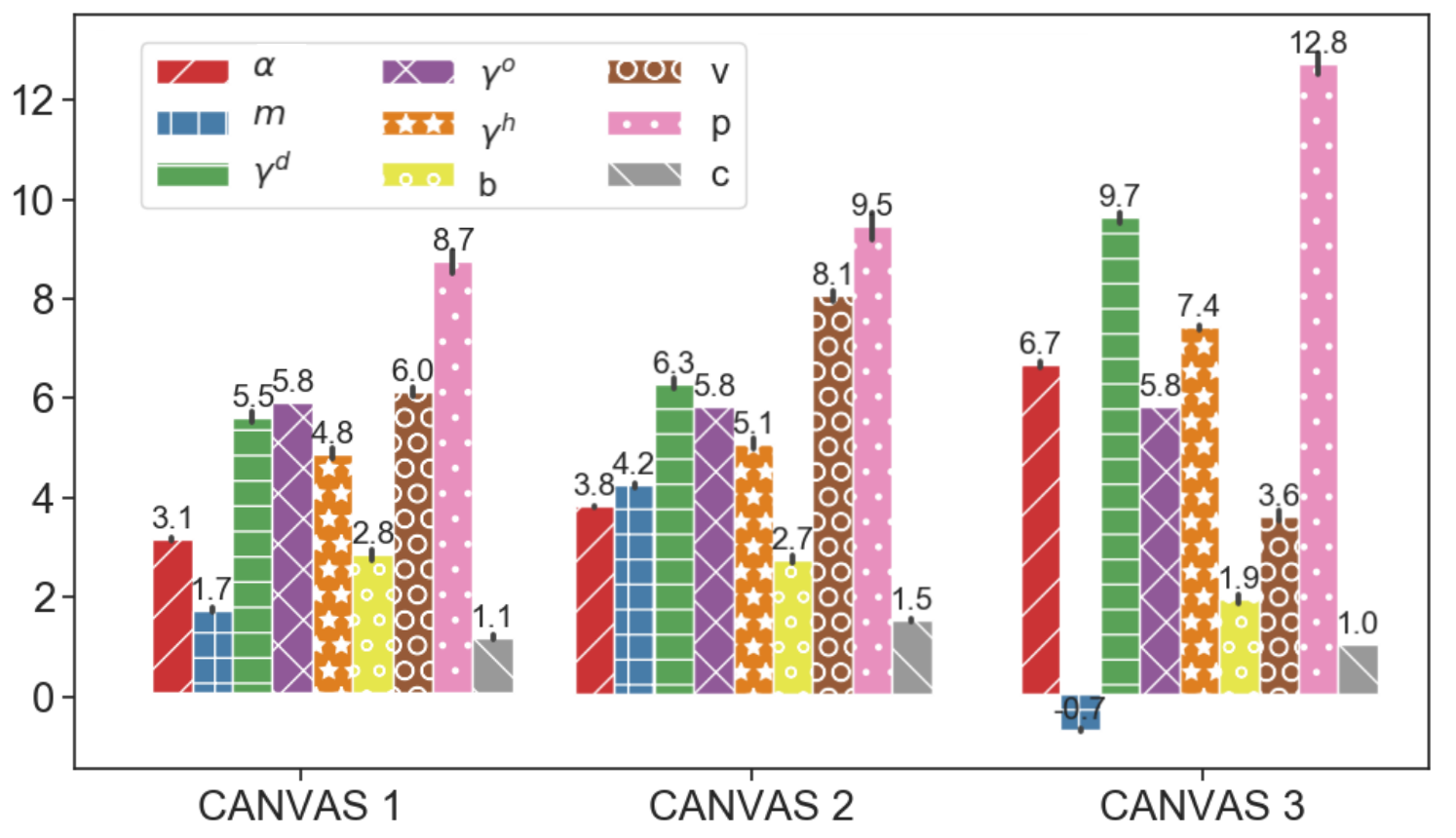}
\caption{Clusters of student learning dynamics characterized by SSHP in CANVAS.}    
\label{fig:pars-canvas}
\end{figure}

\begin{figure}
    \centering
    \includegraphics[width=\linewidth]{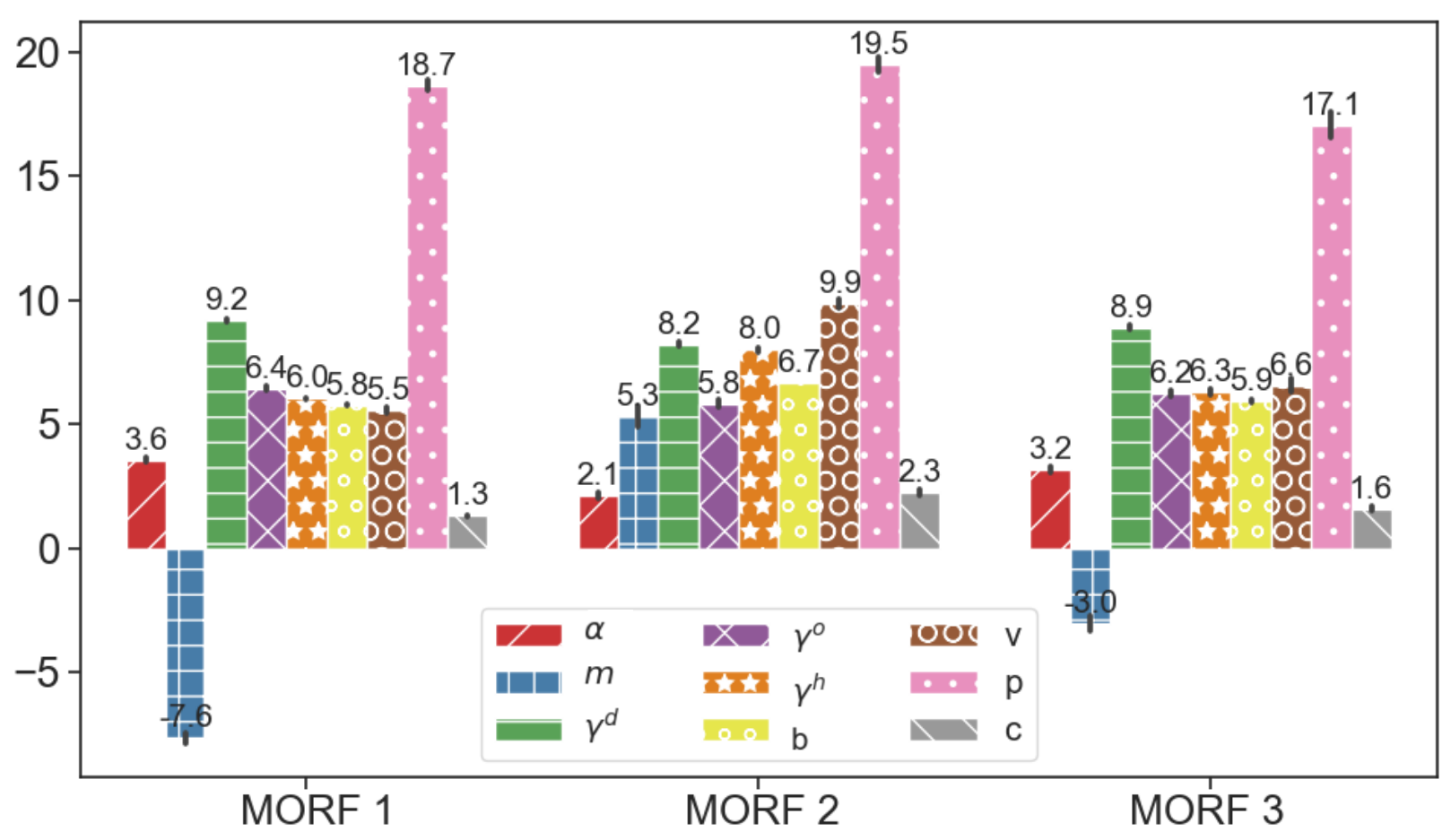}
\caption{Clusters of student learning dynamics characterized by SSHP in MORF.}    
\label{fig:pars-morf}
\end{figure}

To find the optimal number of clusters, we use the elbow method on clustering loss.
In both CANVAS and MORF datasets, the achieved optimal cluster number is $3$, which means that $3$ student-assignment interaction patterns are uncovered in both datasets. 
Figures~\ref{fig:pars-canvas} and~\ref{fig:pars-morf} show the parameter values for cluster centers in CANVAS and MORF datasets, respectively.
For a clearer presentation, $m$ is scaled down by $24$ (time unit changes from hours to days) and $\alpha$ and $b$ are scaled up by $10$ in the figures respectively.
Error bars show the $95\%$ confidence interval within each cluster.

Specifically, by comparing CANVAS $3$ (cluster $3$ in CANVAS dataset) with clusters $1$ and $2$ in CANVAS, 
we can see that the interactions between students and assignments in CANVAS $3$ are shown to be less sensitive to the deadline until much later, when it is too close to the deadline (smaller $v$ and larger $\gamma^d$). 
Also, negative $m$ in CANVAS $3$ indicates late submissions or other assignment-related activities, after the deadline. 
Not only that, but the burstiness of the events in this cluster is also shown to be higher than other clusters (larger $\alpha$).
One possible explanation is that the students in this cluster procrastinated on the assignments in it and only started to work on them much later than they should have, which explains the bursty and intense activities close to the deadline.
Furthermore, we can see that the effect of assignment opening or availability wears off much faster in CANVAS $3$ (smaller $b$), meaning that the period of time that this cluster is affected by assignment opening is shorter. 
This suggests that overall, this cluster is less sensitive to the assignment opening. 
When it comes to student habit, we see that the peak of periodicity shows up at a later time (large $p$), indicating that the students in CANVAS $3$ interact with the course usually later during the day, comparing with CANVAS $2$ and $1$. 
On the other hand, even though the differences are shown to be smaller when comparing CANVAS $1$ and CANVAS $2$ clusters, many of them are significant. 
Particularly, the results clearly show that CANVAS $2$ is more sensitive to the deadline in the sense that assignment-related activities are finished much earlier (larger positive $m$ and $\gamma^d$).
Their base activities triggered by student habits are also shown to be more intense (higher $c$) even though their peak time is usually later during the day (larger $p$).
So to conclude, learning pattern in CANVAS $3$ suggests procrastinating-like behaviors, with less sensitivity to the deadline and the assignment opening, as well as more bursty and intense behaviors. On the other hand, learning patterns in CANVAS $2$ suggests an ``early birds'' type of learning behavior, in which assignment-related activities are finished earlier by around $4$ days. 
Also, they tend to be more sensitive to the opening of assignment, with less bursty behaviors, which can be interpreted as an opposite behavior of procrastination.

Similarly in MORF (Figure~\ref{fig:pars-morf}), different characteristics are uncovered by the discovered clusters.
We can see that the effect of the deadline starts late and ends late (smaller $v$ and smaller negative $m$) for MORF $1$ and $3$. 
Also, student activities on assignments in MORF $1$ and $3$ are more bursty (larger $\alpha$). 
On the other hand, MORF $2$ is more sensitive to the effect of assignment opening for a longer period of time (larger $b$), and student habits also seem to have a stronger effect on MORF $2$, suggested by larger $c$ and $\gamma^h$.
Overall, we can conclude that MORF $2$ activity patterns represent the ``early birds'' type and MORF $1$ activity patterns show the most procrastination-like behaviors among the 3 clusters. 

By comparing the clusters where procrastination-like behaviors are suggested between the two datasets, 
we can see that the parameters show different strategies in them. 
Specifically, in MORF $1$, less bursty and more delayed submissions are observed (smaller negative $m$ and smaller $\alpha$)
than in CANVAS $3$, which can be an indication of procrastination.
Another potential explanation for this difference can be the different nature of the courses as we mentioned in the section of ablation study, where the penalty of late submissions can be stronger in CANVAS than in MORF.

\begin{figure}
    \centering
    \includegraphics[width=\linewidth]{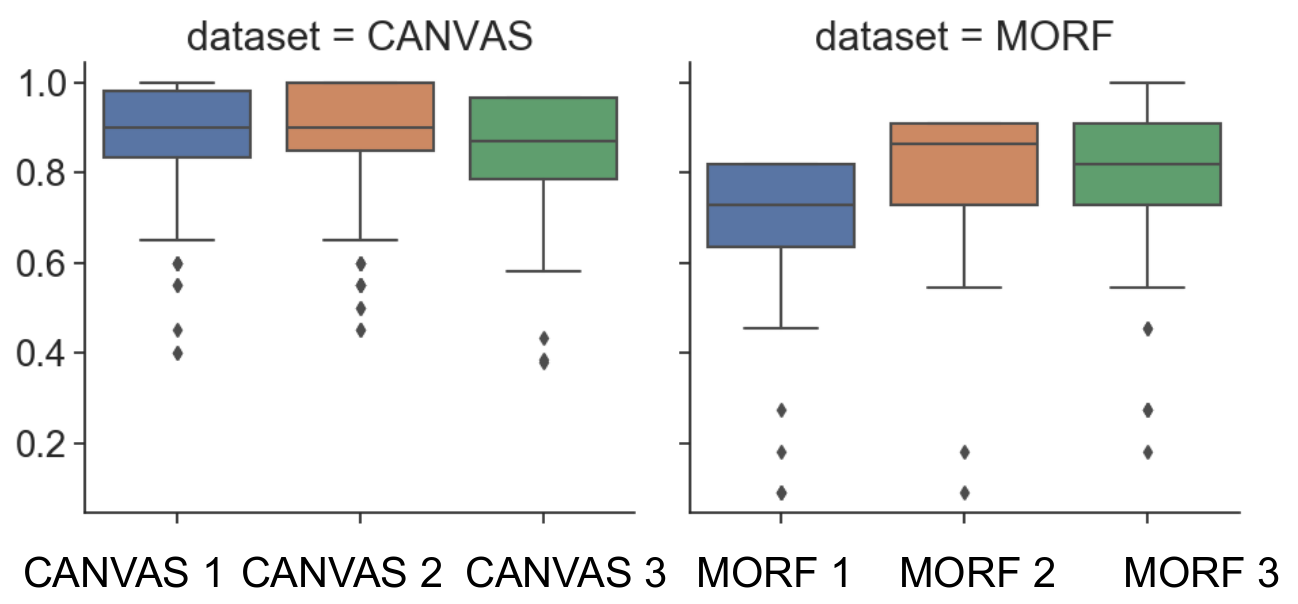}
\caption{Clusters of student learning dynamics characterized by SSHP in CANVAS and MORF.}    
\label{fig:grades}
\end{figure}

\subsection{Association with Grades}
To show the association between student activity patterns on assignments and their performance in them, we check the student grades on assignments in each cluster.
The results are presented as box plots shown in Fig.~\ref{fig:grades}.
As we can see, median grades in CANVAS $3$ and MORF $1$ are visibly smaller than other clusters in their datasets. 
But also, the distribution of grades in each two clusters are different.
To see if the differences of grade distributions between clusters are significant, for each of the datasets, we conduct a Kruskal-Wallis test on the grades between any two clusters discovered by SSHP. 
We find out that all the p-values are significantly smaller than $0.05$, suggesting significant differences in the grade distribution between all clusters. 
Combining these observations with the conclusions from Figures~\ref{fig:pars-canvas} and ~\ref{fig:pars-morf}, we see that clusters that show procrastination behaviors with less sensitivity to the deadlines and assignment openings (CANVAS $3$ and MORF $1$) also are shown to have significantly lower grades.
We can conclude that clusters with more procrastination-like behaviors are associated with lower grades in both datasets.
This demonstrates that SSHP can capture underlying student activity patterns with meaningful parameters that can be used as good indicators of procrastination behaviors and student performances. 

\section{Conclusion}
\label{sec:conclusion}
In this work, we proposed a novel stimuli-sensitive Hawks process model (SSHP) to represent student's cramming and procrastination behaviors in online courses, according to their activities.
Our model captures three types of external stimuli in addition to the internal stimuli between activities, i.e., the effect of assignment deadline, assignment availability, and student's personal habits. 
SSHP models all student-assignment pairs jointly, which enables the model to generate personalized predictions for both partially missing and completely missing activity sequences.
Our experiments on both synthetic and real-world datasets demonstrated SSHP's superior performance comparing to the state-of-the-art baseline approaches, especially in the more challenging task of future time prediction for time sequences where the history is completely missing.
Our ablation studies on SSHP showed that each component of our model is necessary for achieving its superior performance.
Finally, we demonstrated that not only SSHP excels at future time predictions, but also its model parameterization provides meaningful interpretations and insights into the association between students' procrastination patterns and their grades.
Particularly, we discovered $3$ clusters of behaviors on assignments: one with stronger procrastinating behaviors, with less sensitivity to the deadline and the assignment opening, as well as more bursty and intense behaviors; another one with ``early birds'' type of learning behaviors, with more sensitivity to deadlines and less bursty behaviors; and a third one in between the two.
We showed that grade distributions in these clusters have meaningful differences, with the lowest grades associated with procrastinating-like behaviors.

\bibliographystyle{ACM-Reference-Format}
\bibliography{refs}

\end{document}